\documentclass[journal]{IEEEtran}
\usepackage{cite,amsmath,amssymb,graphicx,booktabs,multirow,array,url}
\usepackage[table]{xcolor}
\usepackage[hidelinks]{hyperref}
\usepackage{amsmath}
\usepackage{amssymb}
\usepackage{algorithm}
\usepackage{algpseudocode}
\definecolor{resultgreen}{HTML}{E8F2E3}
\renewcommand{\arraystretch}{1.08}
\newcommand{\R}{\mathcal{R}}
\title{Reference-Efficient Multimodal Bitstream-Corrupted Video Recovery via Depth-Conditioned Adaptation}

\author{
Tianyi Liu, Jiaojiao Wang, Ye Lu, 
Kejun Wu,~\IEEEmembership{Senior Member,~IEEE},
Chengtao~Cai,~\IEEEmembership{Senior Member,~IEEE},
Zhengguo~Li,~\IEEEmembership{Fellow,~IEEE},
and Chia-Wen~Lin,~\IEEEmembership{Fellow,~IEEE}
  
\thanks{Tianyi Liu, Jiaojiao Wang, and Ye Lu are with Nanyang Technological University, Singapore. (Email: \{liut0038, jiaojiao001, lu0001ye\}@e.ntu.edu.sg)}
\thanks{Kejun Wu is with the School of Electronic Information and Communications, Huazhong University of Science and Technology, Wuhan 430074, China. (Email: kjwu@hust.edu.cn)}%
\thanks{Chengtao~Cai is with College of Intelligent Systems Science and Engineering, Harbin Engineering University, Harbin 150001, China.}  
\thanks{Zhengguo~Li is with VI Department, Institute for Infocomm Research, Agency for Science, Technology and Research (A*STAR), Singapore.}
\thanks{Chia-Wen Lin is with the Department of Electrical Engineering and the Institute of Communications Engineering, National Tsing Hua University, Hsinchu 30013, Taiwan.}
\thanks{Corresponding author: Kejun~Wu, Chengtao~Cai.}
}

\begin{document}
\maketitle
\begin{abstract}
Recovering visually faithful content from damaged video bitstreams is important for reliable multimedia communication and storage. 
Existing methods exploit local residuals and non-local RGB references of the input video sequence, but both sources can become unreliable after decoding, while processing additional references increases receiver computation and memory. 
This paper studies multimodal bitstream-corrupted video recovery (MBSCVR) using RGB and depth information, where pre-transmission depth-based geometry aids reconstruction when the received appearance is corrupted. 
We propose a novel MBSCVR framework that selects and effectively utilizes a single selected non-local RGB reference per recovery window to achieve higher performance. 
It integrates a reliability-aware multimodal selector that combines valid appearance with depth compatibility before reference encoding. 
Depth-Conditioned Reference Adaptation (DCRA) introduces selected RGB evidence into feature completion and content generation, while frame-aligned depth conditions completion queries. 
On the complete BSCV benchmark with oracle masks, 
MBSCVR uses one selected reference to achieve 32.30 dB on YouTube-VOS and 28.93 dB on DAVIS, surpassing full-reference B2SCVR by 0.27 and 0.33 dB and improving SSIM on both subsets. 
Matched receiver-forward measurements show 36.8\% lower latency and 40.8\% lower peak memory. 
These results demonstrate how compact geometric guidance and adapted appearance processing improve the quality--resource trade-off of multimedia recovery.
\end{abstract}

\section{Introduction}
\label{sec:intro}

\begin{figure*}[t]
\centering\includegraphics[width=\textwidth]{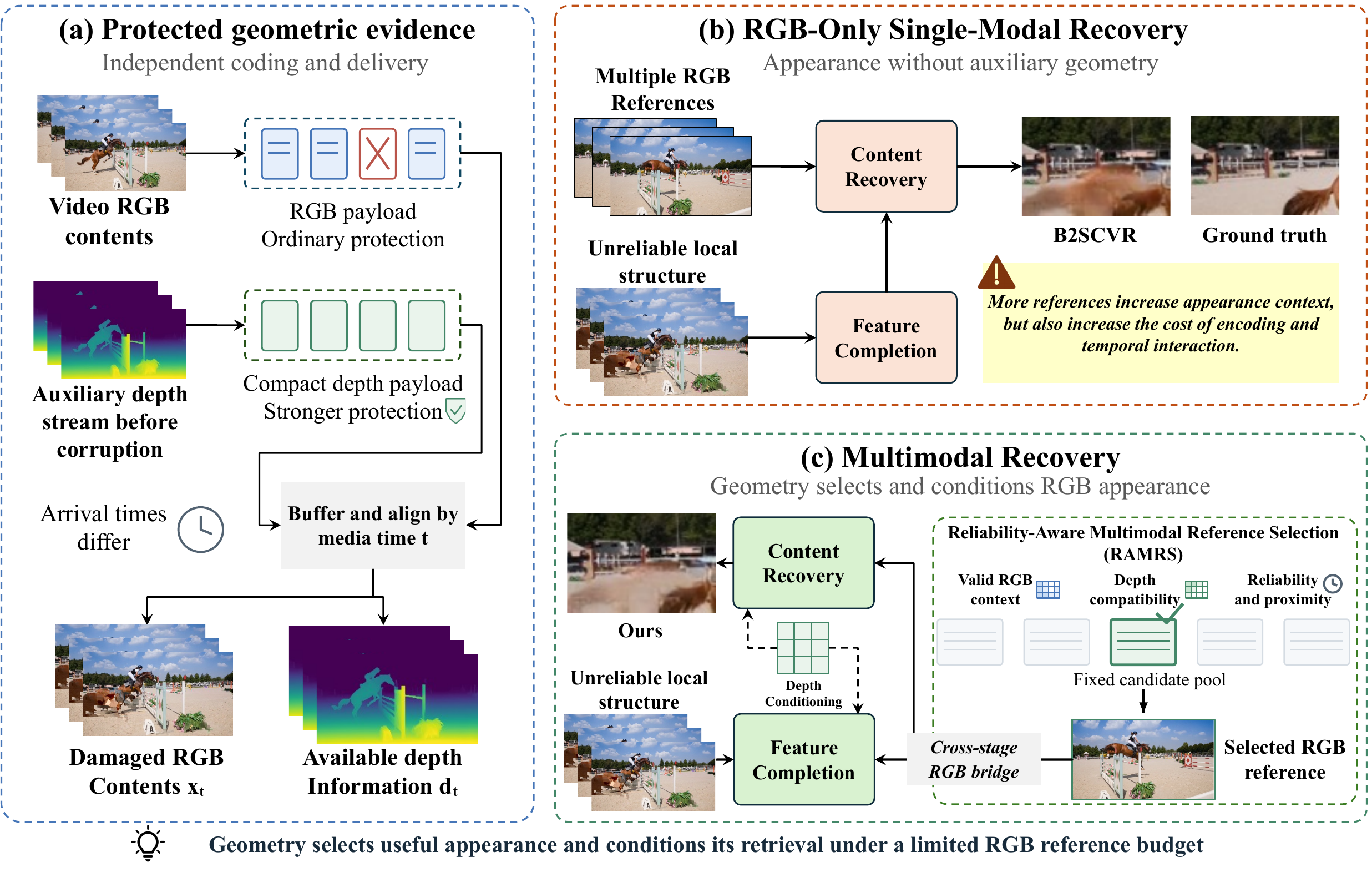}
\vspace{-20pt}
\caption{Motivation and overview of MBSCVR. (a) Auxiliary depth obtained before RGB video corruption is independently coded and delivered with stronger protection, then aligned with the received RGB frames by media time. This provides geometric evidence when local appearance is damaged. (b) RGB-only recovery draws on multiple references to enrich appearance context, increasing reference encoding and temporal interaction costs. (c) MBSCVR uses geometry to select and condition a compact set of RGB references. RAMRS selects relevant appearance using valid RGB context, depth compatibility, and reference reliability, while DCRA introduces the selected appearance and geometric guidance into recovery.}
\label{fig:teaser}
\end{figure*}

Reliable video delivery is fundamental to multimedia communication, entertainment, monitoring, and machine perception. However, compressed video remains vulnerable to transmission and storage errors. Because predictive coding introduces dependencies within and across frames, damage to a small portion of the bitstream can produce extensive and irregular degradation after decoding~\cite{liu2023bitstream, zou2026ntire}. The resulting video often contains intact observations, surviving textures and colors, and misleading artifacts simultaneously. When transport-level repair is unavailable or cannot be completed within the service's buffering allowance~\cite{rfc4588}, receiver-side recovery becomes important for restoring useful visual content and maintaining the quality of the multimedia service.

Bitstream-corrupted video recovery has recently emerged to address this challenging mixture of preserved and damaged information. BSCVR demonstrates that residual texture, color, and structure inside corrupted regions can support feature completion instead of being discarded as in conventional inpainting~\cite{liu2023bitstream}. B2SCVR further improves recovery through automatic corruption localization and corruption-aware residual processing~\cite{liu2025b2scvr}. Both methods also exploit non-local RGB references to supply missing appearance. Nevertheless, their recovery evidence remains centered on the decoded visual stream: when corruption obscures the target structure, the model must identify useful reference content and infer its relationship to the target from incomplete RGB observations. Increasing the number of references expands the available context, but also increases reference encoding, feature storage, and temporal interaction. This quality--resource trade-off is particularly relevant to memory-constrained receivers and recovery services processing multiple videos.

Multimodal video systems offer an additional source of recovery evidence through separately represented scene geometry. 
ISO/IEC 23002-3 explicitly defines depth maps as a type of auxiliary video supplementing a primary video sequence~\cite{iso23002_3_2007}. This representation can be combined with differentiated delivery: RTP supports separate media sessions, while the FEC Framework permits independently configured protection instances for identified source flows~\cite{rfc3550,rfc6363}. Unequal-level protection further allows the sender to vary the protection strength and extent of RTP media data~\cite{rfc5109}. These mechanisms motivate a practical allocation in which a compact depth stream receives stronger protection than the high-rate RGB video stream. Applying additional redundancy to the smaller auxiliary payload requires fewer absolute bytes than applying the same redundancy fraction to the appearance stream, making geometric side information a plausible recovery resource when some RGB content remains damaged.
Motivated by this capability, we consider a \emph{protected-depth auxiliary-stream setting}: RGB and depth are independently coded and packetized, the depth video stream is assigned stronger delivery protection, and available depth is aligned with the damaged RGB video on the media timeline before recovery. Our implementation investigates the reconstruction stage of this setting using depth generated from pristine RGB before corruption. The auxiliary signal thus preserves a description of scene geometry that need not be inferred from the damaged receiver-side appearance.

This setting introduces a useful asymmetry between the modalities. RGB carries the colors and textures needed for faithful reconstruction, whereas depth describes scene layout and geometric structure. The value of depth is therefore not to replace missing appearance, but to help determine which RGB evidence is useful and how it should contribute to recovery. This observation leads to our central question: \emph{can protected geometric side information make one or a few RGB references sufficiently informative for high-quality bitstream-corrupted video recovery?} As illustrated in Fig.~\ref{fig:teaser}, we approach this question by improving the selection and utilization of a limited reference budget, rather than relying on a larger collection.

In this setting, two challenges must be addressed. First, useful references must be identified despite corruption in both the target and candidate RGB frames. A nearby frame may be unreliable, while a visually similar frame may depict incompatible geometry. Reference selection should therefore combine reliable appearance evidence with geometric compatibility and temporal context. Second, selecting an informative frame does not ensure that its appearance will be effectively consumed. Under a small reference budget, the completion stage has less redundant context to compensate for ineffective integration. Reference information must reach the stages where damaged features are completed, and the pretrained completion subsystem must learn to interpret the new geometric and appearance cues together.

To this end, we propose MBSCVR, a reference-efficient multimodal framework built around a \emph{Reliability-Aware Multimodal Reference Selector} (RAMRS) and \emph{Depth-Conditioned Reference Adaptation} (DCRA). RAMRS is a training-free selector combines reference reliability, valid-context RGB similarity, depth compatibility, and temporal proximity to choose a fixed number of references before their encoding by the recovery backbone. DCRA then introduces selected RGB appearance through a cross-stage reference bridge, producing supported residuals for feature completion and subsequent content generation. 
Together, these mechanisms allow geometry to guide the use of reference appearance both when references are selected and when their features are integrated.

Our main contributions are summarized as follows.
\begin{itemize}
    \item To our knowledge, we present the first study of RGB--depth multimodal bitstream-corrupted video recovery. We formulate a reference-budgeted reconstruction framework, MBSCVR, with protected geometric information.

    \item We develop a training-free reliability-aware multimodal reference selector, RAMRS, to retrieve informative reference frames to support both efficient and robust recovery.
    
    \item We propose DCRA to connect selected RGB appearance with feature completion and content generation through a supported residual bridge, while frame-aligned depth conditions completion queries and RGB retains its appearance-bearing role.

    \item We demonstrate that a single selected RGB high-quality reference could surpass SOTA multi-reference methods on the BSCV benchmark. Component studies and matched receiver profiling further characterize the resulting quality--resource trade-off.
\end{itemize}

\section{Related Work}
\subsection{Bitstream-Corrupted Recovery and Video Inpainting}
Video inpainting synthesizes content inside a supplied missing-region mask. STTN learns joint spatial-temporal transformations~\cite{zeng2020learning}, while FuseFormer improves fine-grained interaction through overlapping feature tokenization~\cite{liu2021fuseformer}. E2FGVI combines flow completion, propagation, and content hallucination in an end-to-end framework~\cite{li2022towards}. ProPainter couples dual-domain propagation with a mask-guided sparse Transformer~\cite{zhou2023propainter}. 
These approaches demonstrate the usefulness of temporal context.
Complementarily, HINT~\cite{chen2024hint} preserves visible information through mask-aware encoding for image inpainting. For bitstream-corrupted videos, exploiting available information further requires distinguishing informative residual content from misleading corruption artifacts.
% ~\cite{wu2023endtoend, wu2022gaussian, wu2023highdimensional, wu2024multifocal}
Coding provides compact representation for various visual signals~\cite{joint2003draft, wu2022gaussian, wu2023highdimensional, wu2023endtoend,  wu2024multifocal}, but also introduces error propagation issue when signal is corrupted~\cite{liu2023bitstream}. 
BSCVR specifically exploits surviving structure, texture, and color within corrupted regions~\cite{liu2023bitstream}. B2SCVR adds a corruption detector and foundation-model-guided feature completion~\cite{liu2025b2scvr}. Our work retains this recovery backbone and addresses a different bottleneck: allocating and consuming a limited non-local RGB context. Oracle-mask evaluation isolates this recovery problem from variation in corruption localization.
Metadata-guided diffusion uses motion vectors and frame types to condition blind recovery~\cite{wang2026metadata}. These coding descriptors describe the encoded RGB stream. Our auxiliary depth represents scene geometry generated before corruption and carried separately, enabling a study of complementary visual modalities and their receiver cost.

Efficient temporal modeling is also an active direction in video restoration, as exemplified by the frequency-separable temporal network for video denoising~\cite{tao2025frequency}.
In diffusion-based video editing, HetCache~\cite{liu2026hetcache} reduces redundant computation through heterogeneous caching across denoising timesteps and selective processing of context tokens.
These studies share the goal of reducing computation while preserving visual quality.
Our work pursues this goal by reducing the number of non-local RGB references processed by the recovery backbone while improving their selection and utilization.
In BSCVR and inpainting research, reference-based reconstruction is important and involves two decisions: which evidence to make available and how the model integrates it. BSCVR and E2FGVI leverage standard concatenation~\cite{liu2023bitstream,li2022towards} to achieve reference feature fusion. Sparse attention can reduce interactions after features are encoded, as demonstrated by ProPainter~\cite{zhou2023propainter}. Our selector instead acts before expensive reference encoding by the recovery backbone. It still examines inexpensive descriptors from the candidate pool; its cost depends on video length and belongs in an end-to-end deployment analysis.
Feature transport also need not be a literal replacement. Propagation and attention integrate target and reference context jointly~\cite{li2022towards,zhou2023propainter}. Our bridge follows this continuous-adaptation principle. Selected RGB features enter residual completion and generation pathways instead of overwriting the target with a presumed correct correspondence.

\vspace{-10pt}
\subsection{Multimodal Video Restoration}
\label{sec:related_multimodal}

Multimodal video restoration exploits complementary observations to recover visual information that is difficult to infer from degraded RGB frames alone.
Event cameras provide high-temporal-resolution brightness changes, making them particularly useful for resolving motion obscured by exposure integration.
Kim \textit{et al.}~\cite{kim2022event} address video deblurring under unknown and varying exposure times through exposure-aware event selection and cross-modal feature fusion.
Their subsequent work~\cite{Kim_2024_CVPR} introduces frequency-aware feature enhancement and event-guided video alignment to exploit complementary event information together with temporal dependencies in video frames.
Beyond motion cues, near-infrared (NIR) observations provide an additional spectral source for restoring poorly illuminated visible content.
Niu \textit{et al.}~\cite{Niu_2023_ICCV} investigate NIR-assisted low-light video enhancement using unpaired daytime and nighttime recordings, combining synthesized nighttime training data with bidirectional temporal feature fusion.
These studies illustrate how auxiliary observations can compensate for degradation-specific information loss, with modality-aware selection and temporal integration playing important roles in reconstruction.

Alongside independently captured modalities, estimated depth has been explored as an auxiliary structural representation for video restoration.
Li \textit{et al.}~\cite{li2024depth} propose depth-guided deep video inpainting, which first completes estimated depth and then uses the recovered geometry to guide feature propagation across missing regions.
By distinguishing content at different depth layers, their framework addresses structural ambiguity around foreground--background boundaries and further enhances temporal coherence and texture quality.
For video focal deblurring, Morris \textit{et al.}~\cite{Morris_2025_WACV} introduce depth- and blur-informed Transformers together with image propagation to account for spatially varying defocus.
For binocular video reflection removal, He et al.~\cite{he2025rethinking}
learn transmission-layer depth and combine depth-guided, cross-view, and cross-frame feature enhancement through gated interactions, using geometry to support reflection separation.
Together, these methods motivate a complementary role for geometry: guiding the organization and transfer of visual features while RGB supplies appearance information.
Our work develops this principle for bitstream-corrupted video recovery under a limited RGB reference budget.
Specifically, an auxiliary depth stream guides both reference selection and frame-aligned feature completion, connecting geometric compatibility with the effective utilization of a small set of RGB references.

\section{Methodology}

\begin{figure*}[t]
\centering\includegraphics[width=\textwidth]{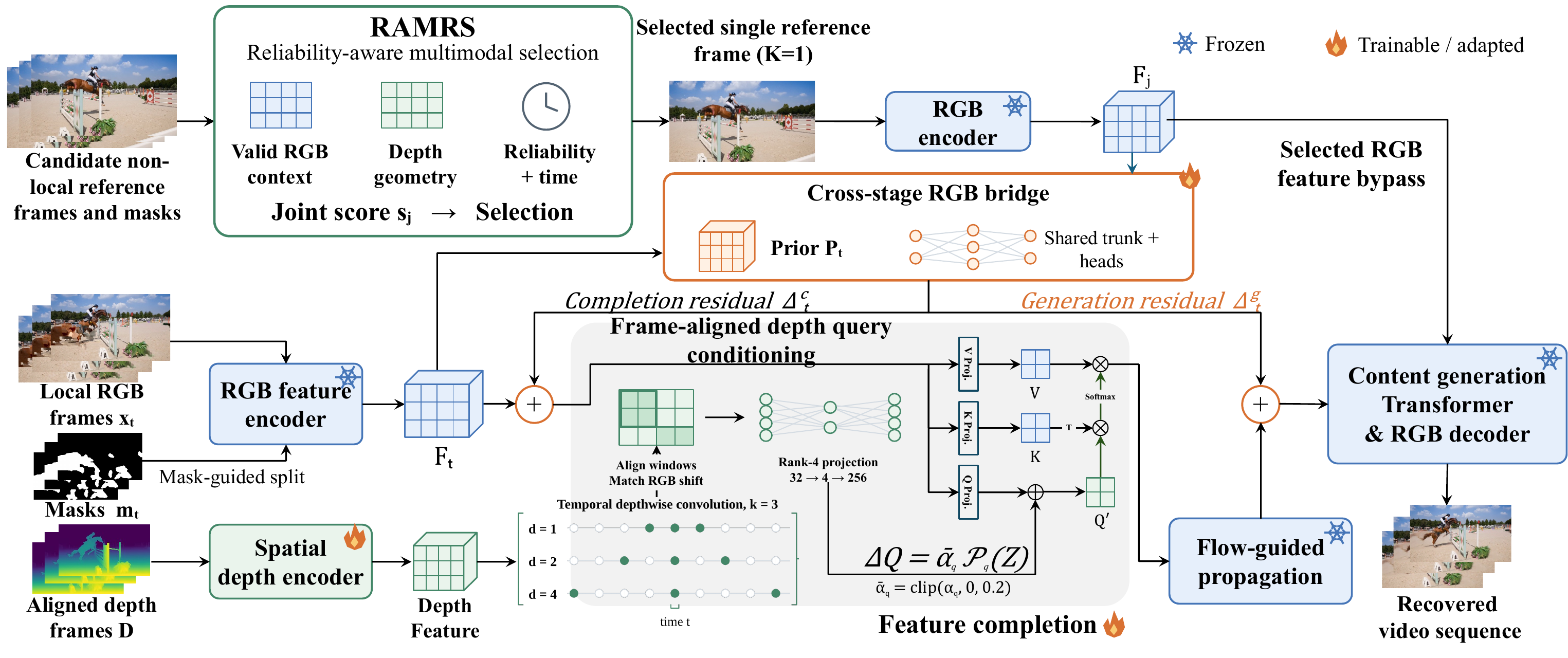}
\vspace{-20pt}
\caption{Architecture of MBSCVR, illustrated with a single selected RGB reference ($K=1$). RAMRS ranks candidates before reference encoding. DCRA integrates the selected RGB features through a cross-stage bridge and conditions feature completion through frame-aligned depth queries. The bridge produces a completion residual $\Delta_t^c$ before feature completion and a generation residual $\Delta_t^g$ after flow-guided propagation. Selected RGB features also retain their direct path to content generation. The depth branch preserves temporal alignment and matches the RGB attention windows, producing $\Delta Q=\overline{\alpha}_q\mathcal P_q(Z)$ through a rank-four projection. The conditioned query is $Q'=Q(F)+\Delta Q$, while keys and values remain RGB-derived. DCRA interfaces and the completion module are jointly adapted; snowflakes indicate frozen modules. The attention diagram omits the scaling factor and bias term $B$ for clarity.}
\label{fig:framework}
\end{figure*}

\subsection{Problem Setting and Framework Overview}
Let $X=\{x_t\}_{t=1}^{T}$ be the decoded corrupted RGB, $M=\{m_t\}_{t=1}^{T}$ be binary corruption masks that use one to indicate damaged areas, and $D=\{d_t\}_{t=1}^{T}$ be aligned depth. 
A local recovery window $\mathcal L$ has a candidate set $\R$. 
Selected non-local indices $\R_K\subseteq\R$ satisfy $|\R_K|=K$ when $|\R|\ge K$. Reconstruction is
\begin{equation}
 \widehat Y_{\mathcal L}=\mathcal F_{\theta,\phi}(X_{\mathcal L},M_{\mathcal L},X_{\R_K},M_{\R_K},D),
\end{equation}
where $\theta$ denotes retained pretrained components and $\phi$ adapted components. 
The budget concerns non-local RGB references, while the local temporal neighbors remain available.

Fig.~\ref{fig:framework} summarizes the framework. The feature encoder, flow-guided propagation module, content generation, and image decoder are frozen. The reliability-aware multimodal reference selection (RAMRS) controls which reference features should be encoded and integrated into the recovery pipeline. Depth-Conditioned Reference Adaptation (DCRA) changes the interface through controlling which of these features and depth should be leveraged in completion.

The framework allocates distinct yet complementary functions to its inputs. 
The chosen RGB frames provide texture-rich representations, whereas depth contributes geometric conditioning. 
RAMRS dictates which visual appearance from the candidate pool should be encoded, and the cross-stage reference bridge exposes that information prior to both feature completion and content generation for recovery. 
Concurrently, the depth pathway modulates how RGB features are retrieved during completion. 
This design maintains the appearance-focused completion while allowing targeted and efficient multimodal and single-reference reconstruction.

\subsection{Reliability-Aware Multimodal Reference Selection}
An informative reference should provide reliable appearance where the target is damaged and depict compatible scene geometry. These requirements are complementary: RGB similarity identifies relevant textures and colors, while depth helps assess structural compatibility when corruption obscures the target appearance. We therefore introduce a training-free Reliability-Aware Multimodal Reference Selector (RAMRS) that combines appearance, visibility, and geometry to allocate a limited RGB reference budget before reference encoding by the recovery backbone.
Let $\mathcal L$ denote the local recovery window and $\mathcal R$ its non-local candidate set in a video of $T$ frames. RAMRS selects a subset $\mathcal R_K\subseteq\mathcal R$ with at most $K$ references. Each frame has decoded RGB $x_t$, a corruption mask $m_t$, and aligned depth $d_t$, with $m_t(p)=1$ indicating damaged content at location $p$. 
% Selection uses low-resolution descriptors and masks on a common spatial grid.

\subsubsection{Reliable Appearance and Spatial Support}

Because the local window shares a limited reference budget, reference selection should prioritize the frames that need the most external support. Let $\overline m_t$ denote the fraction of damaged content in target frame $t$. 
We assign each target a recovery-demand weight
\begin{equation}
 a_t=\frac{\max(\overline m_t,10^{-3})}
 {\sum_{u\in\mathcal L}\max(\overline m_u,10^{-3})}.
 \label{eq:ramrs_target_weight}
\end{equation}
These weights determine how target--reference compatibility is aggregated across the window, giving greater influence to targets with more extensive damage. 
Candidate reliability is evaluated separately, so references are favored when they provide reliable evidence for the targets that need it most.

For a candidate $j\in\mathcal R$, we consider both its overall reliability and its spatial support for the target. 
Its valid-content ratio is $q_j=1-\overline m_j$. For each target--candidate pair $(t,j)$, let $c_{tj}$ denote the fraction of target-damaged descriptor locations that are valid at the same spatial locations in candidate $j$. The window-level support is then
\begin{equation}
 c_j=\sum_{t\in\mathcal L}a_t c_{tj}.
 \label{eq:ramrs_coverage}
\end{equation}
Together, $q_j$ and $c_j$ characterize the availability of reference observations, both globally and at locations requiring recovery.
We combine this availability prior with appearance and geometric compatibility to identify references that are informative for the target.

We use the intact RGB context to assess whether a candidate presents a similar scene appearance. 
For each target--candidate pair $(t,j)$, we compare normalized RGB descriptors only at spatial locations that are intact in both frames. 
Let $e^{\mathrm{rgb}}_{tj}$ denote their mean absolute difference over these jointly intact locations. 
We define
\begin{equation}
 \rho_{tj}=\operatorname{clip}(1-e^{\mathrm{rgb}}_{tj},0,1),
\end{equation}
so that closer agreement in the observed appearance yields a higher similarity score. 
The window-level appearance score is
\begin{equation}
 r_j=\sum_{t\in\mathcal L}a_t\rho_{tj}.
 \label{eq:ramrs_appearance}
\end{equation}
This comparison uses the intact portions of the frames to guide reference selection while excluding corrupted observations.

\subsubsection{Geometric Compatibility and Joint Ranking}
RGB similarity assesses appearance through the intact portions of the target and reference. 
Aligned depth provides complementary geometric evidence at locations where the target RGB is damaged.
We use this evidence to assess whether a candidate has a compatible scene layout and structural boundaries in the regions requiring recovery.

After per-clip depth normalization, we compare each target--candidate pair $(t,j)$ using three complementary measurements. 
Let $e^d_{tj}$ denote the normalized difference in depth values and $e^{\nabla d}_{tj}$ the normalized difference in depth edges, both clipped to $[0,1]$. 
These terms measure agreement in relative scene depth and its spatial transitions, respectively, at target-damaged descriptor locations. 
We additionally use the histogram intersection $h_{tj}$ to characterize similarity in the distribution of depth values.
The pairwise geometric score is
\begin{equation}
 g_{tj}
 = \tfrac12(1-e^d_{tj})
 + \tfrac14(1-e^{\nabla d}_{tj})
 + \tfrac14 h_{tj}.
 \label{eq:ramrs_geometry_pair}
\end{equation}
This combination considers both spatial structure and depth-value distribution. 
As with appearance similarity, we aggregate the pairwise scores according to the window recovery needs:
\begin{equation}
 g_j = \sum_{t\in\mathcal L} a_t g_{tj}.
 \label{eq:ramrs_geometry_window}
\end{equation}

We then combine reference availability with appearance and geometric compatibility. 
The valid-content ratio $q_j$ measures the candidate's overall reliability, while $c_j$ measures the availability of valid observations at target-damaged locations. The scores $r_j$ and $g_j$ assess whether the candidate's appearance and geometry are compatible with joint score
\begin{equation}
 s_j
 = q_j c_j
   \max(r_j,\epsilon)
   \max(g_j,\epsilon)
   \exp\!\left(-\frac{\delta_j}{\max(T-1,1)}\right),
 \label{eq:score}
\end{equation}
where $\epsilon>0$ is a small numerical constant, $T$ is the video length, and $\delta_j$ is the temporal distance of candidate $j$ from the local recovery window. 
The multiplicative formulation encourages references to combine reliable observations with compatible content and the temporal factor favors nearby candidates when the other evidence is comparable.

Candidates are ranked by $s_j$, and the reference budget determines how many are retained. 
We select $K_{\mathrm{eff}}=\min(K,|\mathcal R|)$ candidates, accommodating windows with fewer available references than the requested budget.
For $K=1$, we combine multimodal ranking with the existing temporal sampling rule. 
Let $j_0\in\mathcal R$ be the fixed-stride-sampled reference and $j^*$ the highest-scoring candidate. 
We update the reference from $j_0$ to $j^*$ when
\begin{equation}
 \frac{s_{j^*}-s_{j_0}}
      {\max(|s_{j_0}|,\epsilon_m)}
 \geq \eta,
 \label{eq:ramrs_margin}
\end{equation}
where $\eta>0$ specifies the minimum relative score improvement and $\epsilon_m>0$ stabilizes the denominator. 
This rule allows multimodal evidence to change the temporal choice while retaining it when the scores are similar. 
Algorithm~\ref{alg:ramrs} summarizes the complete selection procedure.

\begin{algorithm}[t]
\caption{Reliability-Aware Multimodal Reference Selection}
\label{alg:ramrs}
\begin{algorithmic}[1]
\Require RGB $X$, corruption masks $M$, aligned depth $D$
\Require Local indices $\mathcal L$, candidate indices $\mathcal R$,
         reference budget $K$, video length $T$
\Require Temporal distances $\{\delta_j\}_{j\in\mathcal R}$;
         fixed-stride reference $j_0\in\mathcal R$ if
         $K=1$ and $\mathcal R\neq\varnothing$
\Require Score constants $\epsilon$, $\epsilon_m$ and update margin $\eta$
\Ensure Selected reference indices $\mathcal R_K$

\State $K_{\mathrm{eff}}\gets\min(K,|\mathcal R|)$
\If{$K_{\mathrm{eff}}=0$}
    \State \Return $\varnothing$
\EndIf

\State Compute RGB and depth descriptors with spatially aligned masks
\State Compute target recovery-demand weights $a_t$
       using~\eqref{eq:ramrs_target_weight}

\For{$j\in\mathcal R$}
    \State $q_j\gets 1-\overline m_j$
    \For{$t\in\mathcal L$}
        \State Compute co-located valid-observation fraction $c_{tj}$
        \State Compute jointly valid RGB similarity $\rho_{tj}$
        \State Compute geometric compatibility $g_{tj}$
               using~\eqref{eq:ramrs_geometry_pair}
    \EndFor
    \State $c_j\gets\sum_{t\in\mathcal L}a_t c_{tj}$
    \State $r_j\gets\sum_{t\in\mathcal L}a_t \rho_{tj}$
    \State $g_j\gets\sum_{t\in\mathcal L}a_t g_{tj}$
    \State Compute candidate score $s_j$ using~\eqref{eq:score}
\EndFor

\State Rank candidates by decreasing $s_j$,
       breaking ties by frame index
\State $\mathcal R_K\gets$ first $K_{\mathrm{eff}}$ ranked candidates

\If{$K=1$}
    \State $j^*\gets$ first ranked candidate
    \If{$(s_{j^*}-s_{j_0})/
          \max(|s_{j_0}|,\epsilon_m)<\eta$}
        \State $\mathcal R_K\gets\{j_0\}$
    \EndIf
\EndIf

\State \Return $\mathcal R_K$ sorted in temporal order
\end{algorithmic}
\end{algorithm}

\subsection{Depth-Conditioned Reference Adaptation}

With a limited reference budget, recovery depends on how effectively the selected appearance is used.
Reference features can inform the completion of damaged local representations as well as subsequent content generation. 
Meanwhile, aligned depth provides structural cues for deciding how RGB features should interact during completion.
Therefore, we propose Depth-Conditioned Reference Adaptation (DCRA) to coordinate these complementary roles: we use a cross-stage bridge that introduces selected RGB appearance into the recovery process, while depth conditions the queries used to retrieve and combine RGB-domain features.

\subsubsection{Cross-Stage RGB Reference Bridge}
The bridge makes selected reference appearance available at two stages: before damaged features are completed and before the completed features
are used for content generation. 
To accommodate differences in reference visibility and reliability, it first constructs a spatially weighted appearance prior.

\begin{figure}[t]
 \centering
 \includegraphics[width=\linewidth]{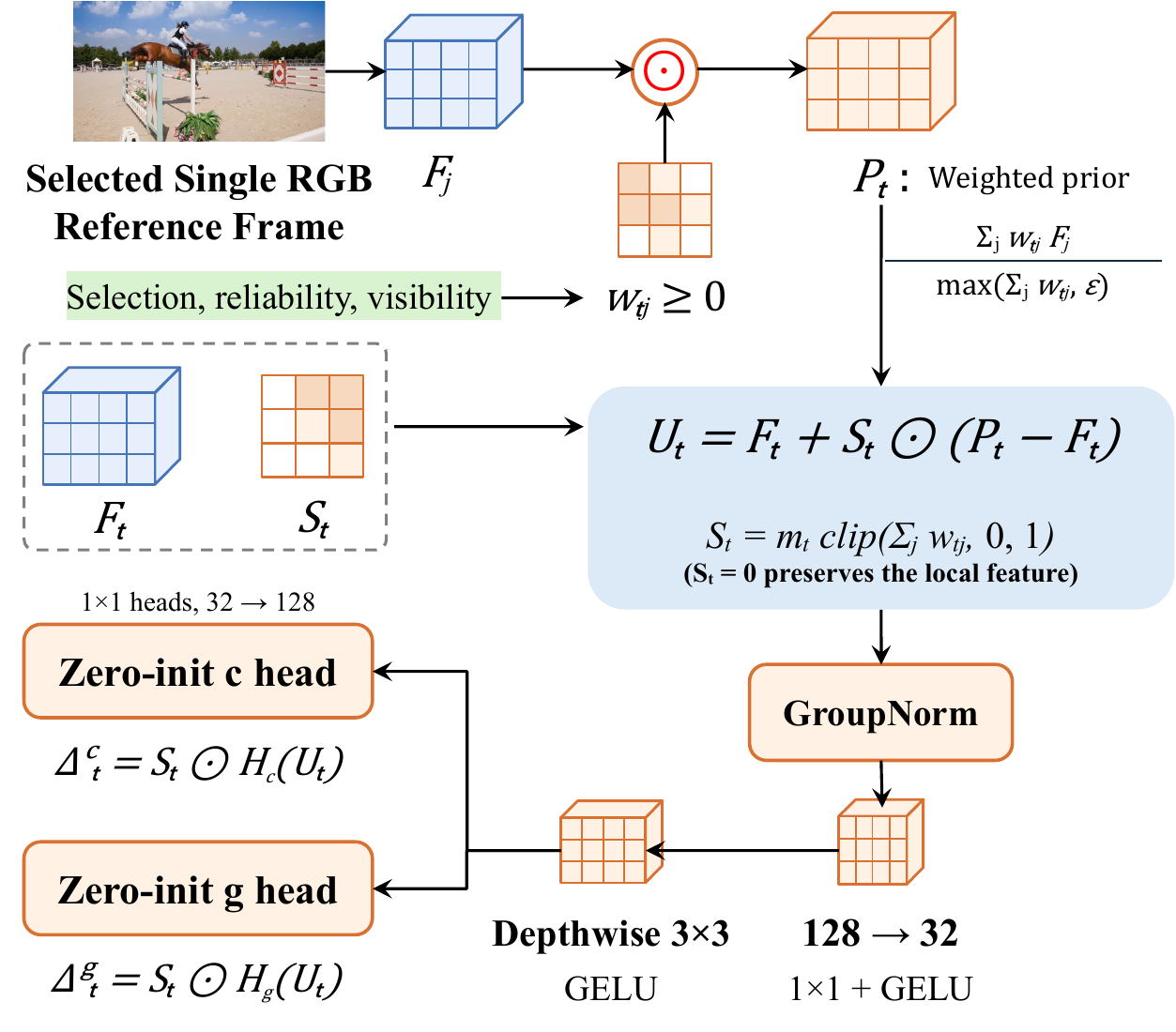}
 \vspace{-20pt}
 \caption{Cross-stage RGB reference bridge within DCRA. Nonnegative spatial weights $w_{tj}$ combine selection evidence, reliability, and visibility to aggregate selected reference features into a normalized RGB prior $P_t$. The support map $S_t$ controls how this prior enters the bridge input $U_t$, concentrating reference guidance on supported damaged regions. A shared lightweight trunk and two zero-initialized output heads produce separate residuals for feature completion and content generation. Here, $H_c$ and $H_g$ each denote the shared trunk followed by the corresponding output head. Summations run over selected references $j\in\mathcal R_K$, and the spatial coordinate $p$ is omitted for clarity.}
 \label{fig:dcra}
\end{figure}

Let $F_t$ be the local RGB feature of target frame $t$, and let $F_j$ be the feature of selected reference $j\in\mathcal R_K$. 
At feature-grid location $p$, we combine the reference selection score, target--reference reliability, and reference visibility into a nonnegative weight $w_{tj}(p)$. 
The reference appearance prior is
\begin{equation}
 P_t(p)=
 \frac{\sum_{j\in\mathcal R_K}w_{tj}(p)F_j(p)}
 {\max\!\left(\sum_{j\in\mathcal R_K}w_{tj}(p),\epsilon\right)},
 \label{eq:dcra_prior}
\end{equation}
where $\epsilon>0$ stabilizes the normalization. 
This co-located aggregation summarizes the selected RGB observations according to their available support.

We then determine where this prior should influence the bridge.
Let $m_t(p)$ denote the target corruption mask at feature resolution, with one indicating damaged content. 
We define the support map as
\begin{equation}
 S_t(p)=m_t(p)\,
 \operatorname{clip}\!\left(
     \sum_{j\in\mathcal R_K}w_{tj}(p),0,1
 \right).
 \label{eq:dcra_support}
\end{equation}
The mask identifies locations requiring repair, while the accumulated reference weights determine the strength of the available guidance.
The bridge input blends the local feature and reference prior as
\begin{equation}
 U_t=F_t+S_t\odot(P_t-F_t),
 \label{eq:dcra_blend}
\end{equation}
where $\odot$ denotes element-wise multiplication. 
A larger $S_t$ gives the reference prior greater influence on this input; when $S_t=0$, the local feature is retained.

The blended representation $U_t$ is used to predict residual corrections for the recovery backbone. 
Let $H_c$ and $H_g$ denote two mappings consisting of a shared lightweight convolutional trunk and separate output heads to produce
\begin{equation}
 \Delta_t^c=S_t\odot H_c(U_t),\qquad
 \Delta_t^g=S_t\odot H_g(U_t).
 \label{eq:dcra_residuals}
\end{equation}
Multiplication by $S_t$ restricts these corrections to supported damaged locations. 
The output heads are zero-initialized, so the bridge starts with zero residual corrections and learns how selected appearance should modify the recovery features.
Within the recovery pipeline, the completion residual $\Delta_t^c$ is added before the feature completion module, allowing reference appearance to inform feature repair itself.
On the other hand, the generation residual $\Delta_t^g$ is added after feature propagation and before temporal generation, providing a separate appearance update for content reconstruction. 
The shared trunk reuses reference evidence, while the two heads learn corrections suited to the different stages.

\subsubsection{Frame-Aligned Depth Query Conditioning}
The reference bridge supplies appearance evidence, but more robust completion should still determine how to combine that evidence with damaged local features if more explicit structural guidance is available. 
Depth contributes to this decision by describing the scene structure corresponding to the target. 
We introduce geometric conditioning through the attention query, which influences the weights assigned to RGB-domain features during completion.

To this end, a spatiotemporal depth encoder captures spatial structure and temporal context while retaining a feature map for each frame.
Its outputs are resized and partitioned to match the RGB completion windows, including their spatial shift. 
This alignment associates each depth token with the corresponding frame and spatial location of the RGB representation.

Let $F$ denote the RGB-domain tokens in a completion attention window and $Z$ the aligned depth tokens. 
The projections $Q(F)$ and $K(F)$ determine the matching scores between RGB tokens, while $V(F)$ provides the feature content aggregated by attention. 
We introduce depth through an additive query correction:
\begin{align}
 \Delta Q
   &=\overline\alpha_q\,\mathcal P_q(Z),
   \label{eq:dcra_query_residual}\\
 Q'
   &=Q(F)+\Delta Q,
   \label{eq:dcra_query}
\end{align}
where $\mathcal P_q$ is a rank-four projection and $\overline\alpha_q=\operatorname{clip}(\alpha_q,0,0.2)$ controls the strength of geometric conditioning.
The resulting completion self-attention is
\begin{equation}
 A=\operatorname{Softmax}\!\left(
      \frac{Q'K(F)^\top}{\sqrt{d_k}}+B
   \right)V(F),
 \label{eq:dcra_attention}
\end{equation}
where $d_k$ is the key dimension and $B$ contains the positional bias and attention-mask terms.

Through $Q'$, depth changes how the RGB features are weighted.  
The keys and values remain RGB-derived, so the content combined for completion continues to come from the appearance representation.
This division connects the two DCRA pathways: the bridge makes selected RGB evidence available to completion, and aligned geometry conditions how completion reads that evidence. 

As the new interfaces change the evidence available to completion, local RGB features now receive selected reference appearance and depth-conditioned attention. 
We therefore focus adaptation on teaching the completion module to effectively integrate these inputs, while
retaining the pretrained stages surrounding it.
Initializing from the released B2SCVR model~\cite{liu2025b2scvr}, we first train the reference bridge and depth-query interfaces with the pretrained backbone fixed, then jointly adapt these interfaces and the feature completion module while the other modules remain frozen.
This allocation concentrates learning on the stage where multimodal evidence is combined to repair damaged representations.
Overall, we optimize
\begin{equation}
 Loss
 =l_{\mathrm{hole},1}
 +0.1l_{\mathrm{valid},1}
 +0.01l_{\mathrm{anchor}},
 \label{eq:completion_objective}
\end{equation}
where $l_{\mathrm{hole},1}$ and $l_{\mathrm{valid},1}$ measure $\ell_1$ reconstruction error in damaged and valid regions, respectively.
The first term emphasizes content recovery, while the second encourages preservation of the observed appearance.
The anchor term averages tensor-wise squared deviations from initialization, regularizing adaptation around the pretrained parameter values.

\section{Experiments}
\subsection{Experimental Settings}
\subsubsection{Benchmark and Evaluation Protocols}
We evaluate MBSCVR on all 661 videos of the BSCV benchmark~\cite{liu2023bitstream}. Recovery is performed at $432\times240$ for both the YouTube-VOS and DAVIS subsets. Following the benchmark protocol, we retain the oracle corruption masks, frame filtering, output composition, and frame-weighted PSNR/SSIM aggregation. LPIPS~\cite{zhang2018unreasonable} and VFID~\cite{kim2019deep} measure complementary image-level and video-feature perceptual distances. The candidate stride is ten, and the default reference budget is $K=1$. The full-reference baseline retains the original B2SCVR reference construction.

Component analyses use all 125 DAVIS videos, with candidate stride five and equal weight for each video. We report corrupted-region PSNR, computed by pooling the squared RGB error over masked pixels within each video before converting it to PSNR. SSIM~\cite{wang2004image} follows the corresponding component evaluator and is also averaged across videos. These analyses share 1,818 recovery windows. For paired comparisons, we estimate percentile 95\% confidence intervals using 10,000 video-bootstrap resamples. The complete benchmark measures overall recovery performance, while the component protocol examines changes in the content requiring repair; their PSNR values therefore have different aggregation scopes.

\subsubsection{Implementation Details}
Depth Anything V2 ViT-S~\cite{yang2024depthanythingv2} estimates sender-side pseudo-depth from pristine RGB frames before corruption. Depth is resized to $216\times120$, normalized per clip, and stored at ten bits. RAMRS operates on compact descriptors before the selected RGB references enter the recovery backbone. RGB similarity requires at least 16 jointly valid descriptor locations; otherwise, its pairwise score is zero. Geometric comparison uses 16-bin depth histograms and evaluates depth and edge differences on target-damaged locations, falling back to the full descriptor grid when fewer than four such locations are available. The minimum target-weight term is $10^{-3}$, the score floor is $10^{-6}$, and the single-reference replacement margin is 0.05, with a denominator floor of $10^{-8}$.

The DCRA bridge uses GroupNorm, a $1\times1$ projection from 128 to 32 channels, and a depthwise $3\times3$ convolution, with GELU after each convolution. Two zero-initialized $1\times1$ heads map the shared features back to 128 channels for completion and generation. The depth encoder produces 32-channel features and uses depthwise temporal convolutions with kernel size three and dilations one, two, and four. Its frame axis is preserved when matching depth features to the RGB attention windows. A $32\!\rightarrow\!4\!\rightarrow\!256$ query projection conditions the first Swin block in the first completion residual group, with its scale constrained to $[0,0.2]$.

We initialize from the released B2SCVR model and first train the DCRA interfaces for 10,000 steps with the pretrained backbone fixed. We then train the interfaces together with the completion module for 10,000 steps, allowing feature completion to accommodate the selected appearance and depth-conditioned queries. Training uses 480 groups constructed from the training data, one group per step, and a fixed single-reference construction; RAMRS is applied at inference. The RGB encoder, SPyNet, TFocal, and image decoder remain frozen. The trainable components comprise the bridge, depth encoder, query adapter, and completion module, totaling 3,611,219 parameters.

For the joint training stage, we use AdamW with weight decay $10^{-5}$ and gradient-norm clipping at one. The bridge and depth/query components use learning rates of $3\times10^{-5}$ and $2\times10^{-4}$, respectively; the enabled completion module parameter groups use $2\times10^{-5}$, $10^{-5}$, and $4\times10^{-6}$. The damaged-region, valid-region, and initialization-anchor loss weights are 1, 0.1, and 0.01. All recovery evaluations use evaluation mode. The network is fixed for the main benchmark, reference-budget and selector comparisons, and inference-time pathway analyses; the training ablation compares training configurations.

\subsection{Quantitative Evaluation}
\begin{table*}[t]\centering
\caption{Complete-test-set BSCV comparison on YouTube-VOS and DAVIS. $^{*}$ denotes BSCV retraining. \textbf{Bold} indicates the best result within the single-reference group.}
\label{tab:main}
\begingroup\fontsize{8}{9.5}\selectfont\setlength{\tabcolsep}{4pt}\renewcommand{\arraystretch}{1.08}
\begin{tabular}{lcccccccccc}\toprule
&&&\multicolumn{4}{c}{YouTube-VOS}&\multicolumn{4}{c}{DAVIS}\\
\cmidrule(lr){4-7}\cmidrule(lr){8-11}
Method&Venue&Refs&PSNR$\uparrow$&SSIM$\uparrow$&LPIPS$\downarrow$&VFID$\downarrow$&PSNR$\uparrow$&SSIM$\uparrow$&LPIPS$\downarrow$&VFID$\downarrow$\\\midrule
STTN~\cite{zeng2020learning}&\multirow{2}{*}{ECCV'20}&\multirow{9}{*}{Multiple}&29.30&0.9150&0.0472&0.0370&26.21&0.8603&0.0634&0.1868\\
STTN$^{*}$~\cite{zeng2020learning}&&&29.94&0.9283&0.0420&0.0350&26.61&0.8731&0.0584&0.1808\\
FuseFormer~\cite{liu2021fuseformer}&\multirow{2}{*}{ICCV'21}&&28.74&0.9010&0.0556&0.0411&26.19&0.8592&0.0662&0.1933\\
FuseFormer$^{*}$~\cite{liu2021fuseformer}&&&29.77&0.9284&0.0421&0.0351&26.71&0.8801&0.0557&0.1761\\
E2FGVI-HQ~\cite{li2022towards}&\multirow{2}{*}{CVPR'22}&&29.69&0.9228&0.0469&0.0363&26.79&0.8765&0.0600&0.1806\\
E2FGVI-HQ$^{*}$~\cite{li2022towards}&&&31.00&0.9473&0.0342&0.0315&27.66&0.9018&0.0491&0.1646\\
ProPainter~\cite{zhou2023propainter}&ICCV'23&&30.40&0.9389&0.0369&0.0497&27.24&0.8920&0.0513&0.1389\\
BSCVR-P$^{*}$~\cite{liu2023bitstream}&NeurIPS'23&&31.56&0.9536&0.0288&0.0296&28.29&0.9147&0.0395&0.1484\\
B2SCVR$^{*}$~\cite{liu2025b2scvr}&ACMMM'25&&32.03&0.9605&0.0279&0.0286&28.60&0.9207&0.0403&0.1461\\
\midrule
B2SCVR&ACMMM'25&\multirow{2}{*}{Single}&31.85&0.9575&0.0300&\textbf{0.0292}&28.49&0.9186&0.0421&0.1485\\
MBSCVR(Ours)&--&&\textbf{32.30}&\textbf{0.9636}&\textbf{0.0288}&0.0295&\textbf{28.93}&\textbf{0.9267}&\textbf{0.0409}&\textbf{0.1465}\\\bottomrule
\end{tabular}\endgroup\end{table*}

Table~\ref{tab:main} presents the complete BSCV comparison. Published results from B2SCVR~\cite{liu2025b2scvr} provide context for STTN, FuseFormer, E2FGVI-HQ, ProPainter, BSCVR-P, and their reported retrained variants. We distinguish these published results from our matched evaluations of MBSCVR and B2SCVR with full and single-reference budgets. Comparisons with MBSCVR use the same oracle-mask setting; published blind recovery results additionally include the effect of automatic corruption localization.

With one selected RGB reference, MBSCVR achieves 32.30 dB on YouTube-VOS and 28.93 dB on DAVIS. Relative to the matched full-reference B2SCVR baseline, PSNR improves by 0.27 and 0.33 dB, while SSIM improves by 0.0031 and 0.0060. At the same single-reference budget, the PSNR gains over B2SCVR are 0.45 and 0.4344 dB. Thus, the complete framework maintains a compact RGB context while improving reconstruction fidelity beyond both reference-budget baselines. This result supports the central premise that the usefulness of reference information depends on both its selection and its integration into recovery.

MBSCVR also improves LPIPS over single-reference B2SCVR on both subsets. Full-reference B2SCVR retains a small advantage in LPIPS and VFID, indicating that access to broader RGB context remains useful for these perceptual measures. The main quantitative advantage of MBSCVR is therefore higher PSNR and SSIM with a single reference, together with better image-level perceptual similarity than the baseline at the same budget.

\subsection{Qualitative Evaluation}
\begin{figure*}[t]
\centering
\includegraphics[width=\textwidth]{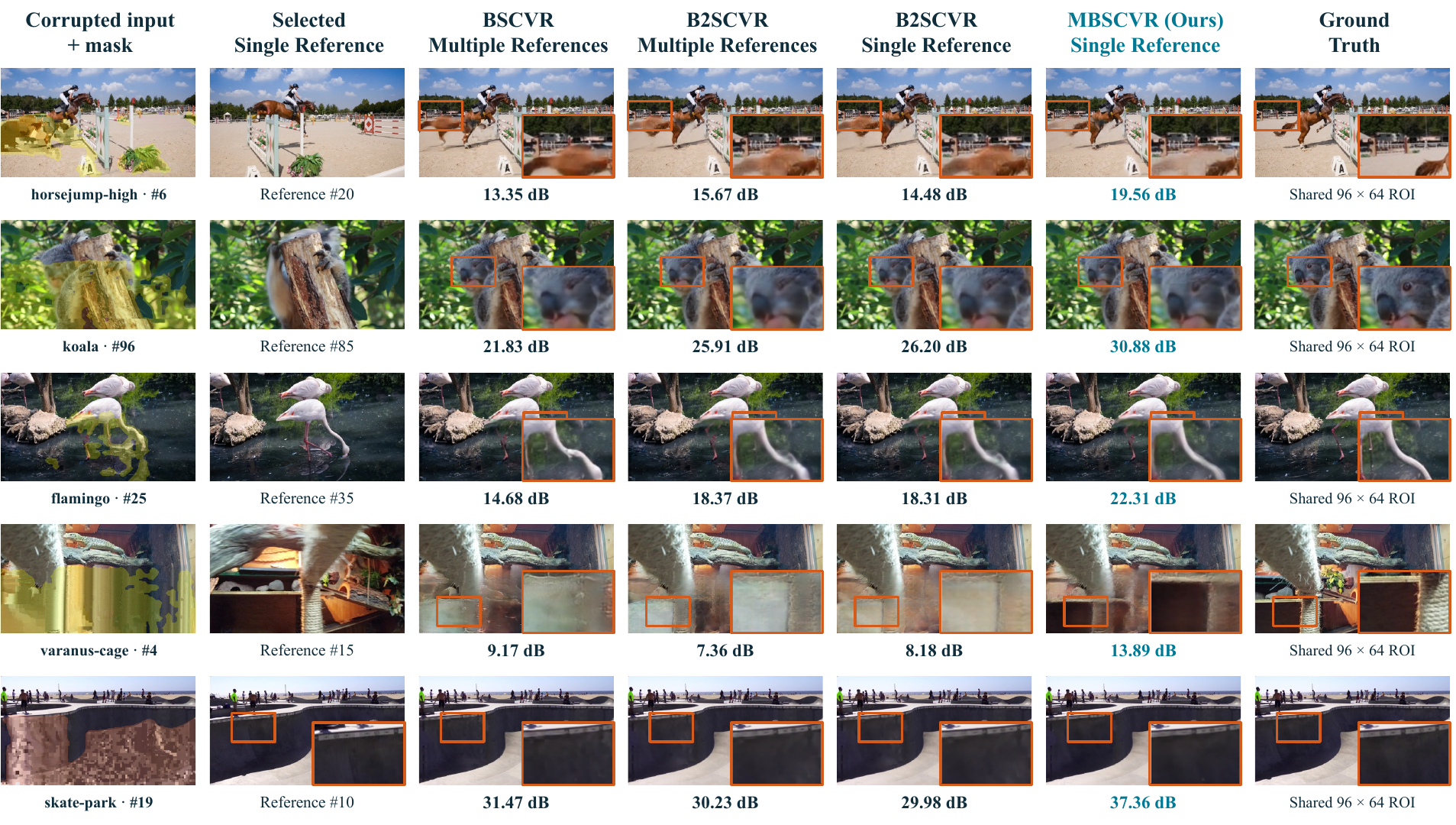}
\vspace{-20pt}
\caption{Qualitative comparison on representative BSCV DAVIS examples. From left to right: corrupted input with the corruption mask overlaid, the RGB reference selected by RAMRS, BSCVR with multiple references, B2SCVR with multiple references, B2SCVR with a single reference, MBSCVR with a single reference, and ground truth. Highlighted boxes and enlarged insets show a common region within each example. MBSCVR combines selected appearance with geometric guidance to recover structural boundaries and local details under a single-reference budget.}
\label{fig:qual}
\end{figure*}
Fig.~\ref{fig:qual} illustrates how selected appearance supports recovery across different types of corruption. In \texttt{horsejump-high}, MBSCVR reconstructs a more coherent boundary between the horse tail and background. In \texttt{koala}, MBSCVR restores a more natural facial appearance, particularly around the eyes, while producing clearer surrounding contours. In \texttt{flamingo}, it reconstructs a more continuous neck structure, whereas the compared methods retain the fragmented structure present in the corrupted input. In \texttt{varanus-cage}, MBSCVR recovers the contours of the supporting frame. In \texttt{skate-park}, the restored wall exhibits colors closer to the ground truth, with residual block artifacts substantially suppressed. These examples illustrate improvements in facial detail, structural continuity, and color consistency.

\subsection{Reference Budget and Receiver Workload}
\subsubsection{Recovery Quality and Reference Budget}
\begin{table}[t]
\centering
\caption{Reference-budget comparison on the 125-video DAVIS component set using the same MBSCVR checkpoint. A budget of three uses all available candidates when fewer than three are present.}
\label{tab:reference_budget}
\begin{tabular}{lrr}
\toprule
Reference budget & Corr. PSNR $\uparrow$ & Corr. SSIM $\uparrow$ \\
\midrule
$K=1$ & 21.3143 & 0.67805 \\
$K=3$ & 21.3797 & 0.68054 \\
\bottomrule
\end{tabular}
\end{table}
Table~\ref{tab:reference_budget} examines the quality available from a larger reference budget. Increasing the budget from one to three raises corrupted-region PSNR from 21.3143 to 21.3797 dB and SSIM from 0.67805 to 0.68054. The paired PSNR gain is 0.0654 dB, with a 95\% interval of $[0.0386,0.0932]$, and 97 of 125 videos improve. The effective average is 2.998 references per window because a few windows supply only two candidates. A larger budget therefore offers a modest additional quality gain, while $K=1$ provides the more economical operating point.

\subsubsection{Receiver Computation}
\begin{table}[t]
\centering
\caption{Receiver-forward measurements on one shared window with six local frames, using float32 on an RTX 3090, two warmups, and ten repeats. Columns report median latency, peak allocated memory, and FLOPs for supported profiler operators. Latency includes device preparation and output transfer.}
\label{tab:runtime}
\begin{tabular}{lrrrr}
\toprule
Configuration & Refs & ms & MiB & TFLOPs \\
\midrule
B2SCVR Full & 9 & 431.7 & 3202.3 & 4.265 \\
B2SCVR $K=1$ & 1 & 270.0 & 1886.6 & 2.405 \\
MBSCVR $K=1$ & 1 & 272.9 & 1895.8 & 2.409 \\
MBSCVR $K=3$ & 3 & 313.4 & 2086.2 & 2.845 \\
\bottomrule
\end{tabular}
\end{table}
The reference budget directly controls the amount of RGB context processed by the recovery network. In Table~\ref{tab:runtime}, single-reference MBSCVR reduces forward latency by 36.8\%, peak allocated memory by 40.8\%, and counted FLOPs by 43.5\% relative to full-reference B2SCVR. Compared with B2SCVR at $K=1$, DCRA adds 2.9 ms and 9.2 MiB. The added conditioning therefore occupies a small part of the recovery workload, allowing the reduction in RGB reference processing to dominate the measured forward cost.
At $K=3$, latency and memory increase to 313.4 ms and 2086.2 MiB, remaining below the full-reference baseline. 
These measurements characterize the recovery forward pass on a shared window. Reference selection, depth decoding, video loading, and metric computation are outside the timed region; sender-side depth estimation and transmission are also outside this receiver-forward measurement.

\subsection{Ablation Study}
We further examine how reference selection and joint interface--completion optimization support recovery with a single RGB reference. 
All comparisons are conducted using the BSCV DAVIS subset.

\begin{table}[t]
\centering
\caption{Reference selection and DCRA optimization on the DAVIS subset. 
Panel A uses the same trained model, candidate pool, and single-reference budget. 
Panel B compares gains from a basic no-DCRA pipeline, with DCRA but interface only variant, and joint interface--completion optimized version.}
\label{tab:selection_integration}
\begin{tabular}{lrr}
\toprule
Configuration & Corr. PSNR $\uparrow$ & Corr. SSIM $\uparrow$ \\
\midrule
\multicolumn{3}{l}{\textit{A. Reference selection}} \\
Fixed stride & 21.1716 & 0.67417 \\
RAMRS & 21.3143 & 0.67805 \\
\midrule
\multicolumn{3}{l}{\textit{B. DCRA optimization}} \\
No DCRA & 20.6456 & 0.65272 \\
Interfaces only & 20.9957 & 0.66107 \\
Interfaces + completion & 21.3143 & 0.67805 \\
\bottomrule
\end{tabular}
\end{table}

\subsubsection{Recovery-Aware Reference Selection}
With a limited reference budget, the selected frame determines which external appearance is available for recovery. RAMRS makes this choice according to candidate reliability and compatibility with the local recovery window. Panel A of Table~\ref{tab:selection_integration} compares this strategy with fixed-stride sampling while keeping the trained model, candidate pool, and reference count unchanged. RAMRS improves corrupted-region PSNR from 21.1716 to 21.3143 dB and SSIM from 0.67417 to 0.67805. The paired PSNR gain is 0.1427 dB, with a 95\% confidence interval of $[0.0841,0.2114]$, and 90 of 125 videos improve. These results support allocating the available reference budget according to the visual evidence relevant to recovery.

\subsubsection{Joint Optimization of Reference Conditioning and Completion}
DCRA introduces selected appearance and geometric guidance into feature completion. Joint optimization allows its conditioning interfaces and the completion features that interpret their outputs to adapt together. Panel B of Table~\ref{tab:selection_integration} evaluates this design. Optimizing only the DCRA interfaces while keeping the backbone frozen achieves 20.9957 dB, compared with 20.6456 dB for the no-DCRA baseline. Jointly optimizing the interfaces and the completion module further improves PSNR to 21.3143 dB. The interface-only and joint variants start from the same DCRA checkpoint and use the same number of additional optimization steps. Joint optimization yields a paired gain of 0.3186 dB over interface-only optimization, with a 95\% confidence interval of $[0.2607,0.3792]$, and 107 of 125 videos improve. This comparison supports jointly adapting reference conditioning and feature completion within DCRA. The RGB encoder, flow estimator, temporal generator, and image decoder retain their pretrained parameters.

The jointly optimized model makes use of the selected RGB evidence. Keeping its parameters fixed, we zero the non-local RGB features and mark them unavailable while retaining the local observations and selected frame indices. Corrupted-region PSNR decreases by 0.5072 dB, with a 95\% interval of $[-0.6474,-0.3852]$, and 117 of 125 videos degrade. This sensitivity confirms that the selected appearance contributes to recovery through the affected reference pathways, including the backbone's direct route to temporal generation.

\section{Conclusion}
We presented MBSCVR for RGB--depth multimodal recovery of bitstream-corrupted video. RAMRS selects a compact set of RGB references according to the local recovery needs, and DCRA integrates their appearance into recovery with depth-conditioned completion queries. Training the DCRA interfaces together with feature completion enables their integration into the pretrained backbone. On the complete BSCV benchmark, a single selected reference improves PSNR and SSIM over full-reference B2SCVR, while the matched receiver-forward evaluation shows lower latency, memory use, and counted FLOPs. These results demonstrate an effective balance between reconstruction fidelity, RGB reference workload, and auxiliary geometric information.
\bibliographystyle{IEEEtran}
\bibliography{references}
\end{document}